\documentclass[10pt,twocolumn,letterpaper]{article}

\usepackage{caption} 
\usepackage{authblk}
\usepackage{wacv}
\usepackage{times}
\usepackage{epsfig}
\usepackage{graphicx}
\usepackage{amsmath}
\usepackage{amssymb}
\usepackage{booktabs}
\usepackage{enumitem}
\usepackage{adjustbox}
\usepackage{placeins}
\usepackage{subfiles}
\usepackage{txfonts}

\usepackage{spconf}

\usepackage[utf8]{inputenc}

\wacvalgorithmstrack   
\wacvfinalcopy 

\usepackage[pagebackref=true,breaklinks=true,colorlinks,bookmarks=false]{hyperref}

\begin{document}

\title{Enhancing Next Active Object-based Egocentric Action Anticipation with Guided Attention}


\name{Sanket Thakur\textsuperscript{1,4}, Cigdem Beyan\textsuperscript{2,1}, Pietro Morerio\textsuperscript{1}, Vittorio Murino\textsuperscript{3,1}, Alessio Del Bue\textsuperscript{1}}
\address{\textsuperscript{1} Pattern Analysis and Computer Vision, Istituto Italiano di Tecnologia, Genoa, Italy \\
\textsuperscript{2} University of Trento, Trento, Italy \\
\textsuperscript{3} University of Verona, Verona, Italy \\
\textsuperscript{4} DITEN, University of Genoa, Genoa, Italy} 
\maketitle
\thispagestyle{empty}


\begin{abstract}
Short-term action anticipation (STA) in first-person videos is a challenging task that involves understanding the next active object interactions and predicting future actions. Existing action anticipation methods have primarily focused on utilizing features extracted from video clips, but often overlooked the importance of objects and their interactions. To this end, we propose a novel approach that applies a guided attention mechanism between the objects, and the spatiotemporal features extracted from video clips, enhancing the motion and contextual information, and further decoding the object-centric and motion-centric information to address the problem of STA in egocentric videos. Our method, GANO (Guided Attention for Next active Objects) is a multi-modal, end-to-end, single transformer-based network. The experimental results performed on the largest egocentric dataset demonstrate that GANO outperforms the existing state-of-the-art methods for the prediction of the next active object label, its bounding box location, the corresponding future action, and the time to contact the object. The ablation study shows the positive contribution of the guided attention mechanism compared to other fusion methods. Moreover, it is possible to improve the next active object location and class label prediction results of GANO by just appending the learnable object tokens with the region of interest embeddings. 
\end{abstract}

\footnote{{\fontencoding{TS1}\selectfont\symbol{"A9}} 2023 IEEE. Personal use of this material is permitted. Permission from IEEE must be obtained for all other uses, in any current or future media, including reprinting/republishing this material for advertising or promotional purposes, creating new collective works, for resale or redistribution to servers or lists, or reuse of any copyrighted component of this work in other works.}
\section{Introduction}
\label{sec:intro}

Short-term action anticipation in egocentric videos is the task of predicting the actions that are likely to be performed by a first-person in the near future, along with foreseeing a next-active-object interaction and an estimate of the time at which the interaction will occur. This is a challenging task due to several factors such as camera motion, occlusions, and the complex and dynamic nature of the environments in which the videos are captured. This problem has several practical applications such as in the domain of augmented reality, where the camera worn by a subject captures that person's actions and interactions with the environment. The computer vision community has gathered significant progress in the field of action anticipation in egocentric videos, which predicts only the action labels \cite{avt,liu2019forecasting,rulstm, memvit2022}. However, the use of the next active objects for anticipating future actions has just emerged in \cite{ego4d}. Based on the description of \cite{ego4d}, the task of short-term anticipation remains challenging since it requires the ability to anticipate both the mode of action and the time at which the action will begin, known as the time to contact.

The next active objects play a crucial role in understanding the nature of interactions happening in a video. They provide important context for predicting future actions as they indicate which objects are likely to be involved in the next action \cite{tpami_contact}. For example, if a person is reaching for a cup on the table, the cup would be considered the next active object and the possible actions matching with a cup can be retrieved by a model. Such an approach reduces the search space of the model, and can even make more accurate predictions. In this vein, we propose a novel approach for addressing the problem of STA in egocentric videos. Our approach utilizes a guided attention mechanism between the spatiotemporal features extracted from video clips and objects to enhance the motion information and separately decode the object-centric and motion-centric information. Our model is a multi-modal, end-to-end, a single transformer-based network and it is called Guided Attention for Next Active Object (GANO).

The main contribution of this paper is to show the importance of the proposed guided attention mechanism for the next active object-based STA. To the best of our knowledge, no prior work has explored the relationship between the next active objects and future actions in the context of egocentric action anticipation. Our approach aims to better capture the visual cues related to the next active objects, which we assume are highly correlated with the action that will follow.
The proposed GANO model is trained and evaluated on the largest egocentric video dataset: Ego4D \cite{ego4d}. Experimental results demonstrate that GANO outperforms the state-of-the-art (SOTA) egocentric action anticipation methods. Additionally, we provide an analysis investigating the impact of guided attention on the performance of the GANO model. The results justify that incorporating guided attention, in other words, combining the information from spatiotemporal features and objects, improves the STA performance with respect to other fusion mechanisms.

\section{Related Work}
\label{sec:related_works}
The prior works focusing on the next active objects and action anticipation in egocentric videos are discussed in this section due to their relevance to the scope of our work. \\

\noindent
\textbf{Action Anticipation in Egocentric Videos.}
The short-term action anticipation in \emph{first-person} videos formalized in \cite{ek55}, has recently gained popularity \cite{liu2019forecasting,rulstm,miech2019leveraging,tpami_contact} perhaps due to its applicability on wearable computing platforms \cite{avt}. Several approaches focus on learning video representations with Convolutional Neural Networks (CNNs) \cite{feichtenhofer2019slowfast,rulstm,miech2019leveraging}, leveraging such as hand movements \cite{liu2019forecasting}, hand-object contact points \cite{liu2019forecasting} while some of them model the activities \cite{tpami_contact} and others aggregate the past contextual features \cite{rulstm,sener} to model interaction and perform predictions.
More recently, researchers have explored the use of Vision Transformers \cite{vit}. Girdhar and Kristen \cite{avt} propose causal modeling of video features, where they introduce a sequence modeling of frame features to decode future interaction in consecutive future frames.
On the other hand, Wu et al. \cite{memvit2022} perform multi-scale representation of frame features by hierarchically attending the previously cached ``memories'', but does not incorporate object-centric features necessary for STA tasks.  In this paper, we utilize the Multi-Scale Vision Transformer (MViT) network \cite{mvit1,mvit2} which introduced multiscale vision feature hierarchy for long-term video modeling,  as the foundation of our architecture to extract and contextualize motion information from a given video clip. To further enhance the performance of the Multi-Scale network in predicting motion-based outputs, we introduce a Guided Network (which is a unique property of our model with respect to the relevant prior art) that enables the network to attend to objects. \\

\noindent
\textbf{Next Active Objects.}
Pirsiavash and Ramanan \cite{ADL} introduced the concepts of \emph{active} and \emph{passive} objects in the context of egocentric vision. The active objects are those the user is interacting with, and the passive objects are the ones in the background. 
Given the definition of \cite{ADL}, Dessalene et al. \cite{tpami_contact} propose a method to detect the \emph{next active objects} by predicting the location of the object that will be contacted by the \emph{hand}. A limitation of that work \cite{tpami_contact} is that it requires the hands and the next active objects to be visible in the frames. 
Furnari et al. \cite{furnari2017next} infers the next active objects with object tracking and such an approach is limited to performing detection only in one next future frame. In other words, their method is not suitable for short-term anticipation tasks where the action could start at any time in the future. 
In contrast to \cite{tpami_contact,furnari2017next}, Thakur et al. \cite{anacto} incorporate object detection by leveraging a transformer model for combined modeling of both RGB and object features. Specifically, that approach \cite{anacto} focuses on anticipating the location of the next active object(s) several frames ahead of the last observed frame, which is crucial for action anticipation. By incorporating object features in the transformer model, the method is able to accurately locate the next active object(s) in future frames.

It is worth mentioning that, the proposed method primarily focuses on predicting future actions and estimating the time at which the interaction with an object starts. However, its capability also involves the prediction of the next active objects' location in terms of the bounding boxes, and the objects' class (called noun).

\begin{figure*}[t!]
\centering
\includegraphics[width=\linewidth, height=0.35\linewidth]{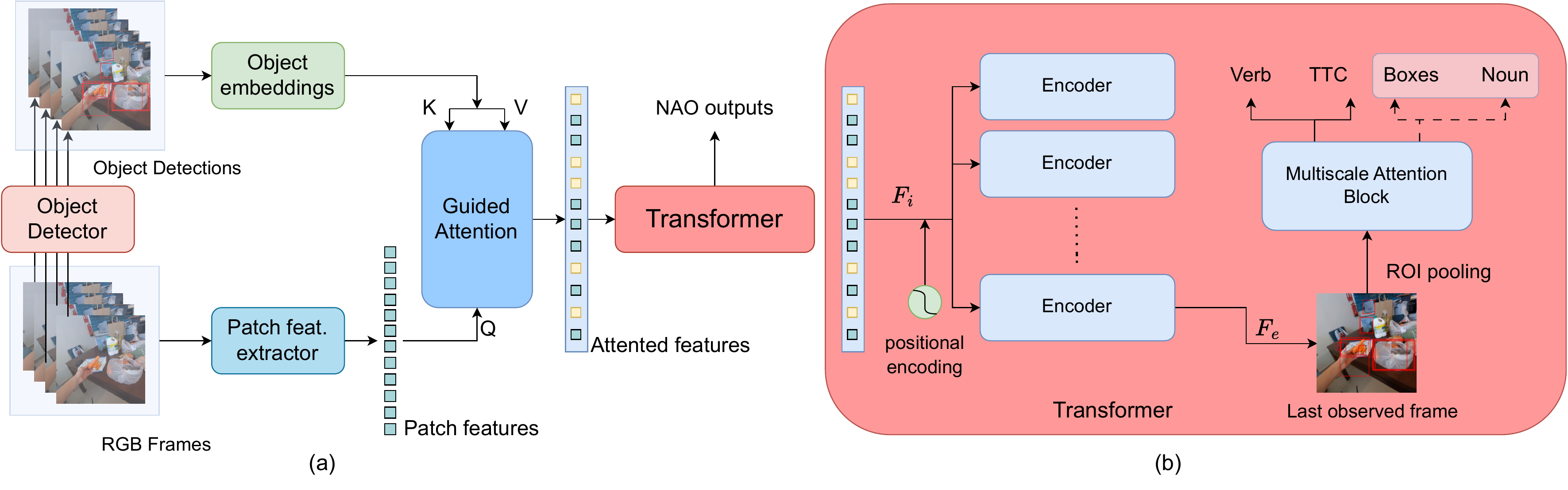}
\caption{Our GANO model uses a 3D convolution layer to extract frame patch features and an object detector to extract object embeddings from corresponding frames. These features are then fused together using a guided attention network to generate attended object-patch features. The attended features, $F_i$, are then given to a transformer encoder layer, along with positional encoding, to obtain features ($F_e$) from the last encoder layer. $F_e$ are then used to extract Regions of Interest (ROIs) from the last observed frame, which are used to predict future actions and Time to Contact (TTC) ($\hat{v}$ and $\delta$, respectively) for the detected objects. Additionally, we append the learnable tokens to the ROI embeddings, creating a fixed query length, and use them to generate the next active object-related predictions.
}
\centering
\label{fig:method}
\vspace{-10pt}
\end{figure*}

\section{Proposed Method}
\label{sec:pagestyle}
Recalling the goal which is to anticipate the mode of action (such as ''pick up" or ''put down), as well as the time at which the action will begin in the future (known as the time to contact ($\delta$)) along with next active object (i.e., the noun class) and its position \textit{wrt} the last observed frame, the model is allowed to process the video sequence $V = \{{v_i\}^T_{i=1}}$, where $v_i \in {R}^{C\times{H_o}\times{W_o}}$ up to frame $T$ where it must locate the position of the next active object in the last observed frame and also anticipate the future interaction with that object in $\delta$ secs, where $\delta$ is unknown. The use of the next active objects is a key aspect of this task since it provides the model to focus on the objects that are likely to be involved in the next action. \\

\noindent
\textbf{Feature extraction.}
The features extracted from an observed video segment $V$ involve (a) the patch features that are extracted from a 3D convolution (Conv3D) layer as in \cite{motionformer,video_swin} and (b) the objects located by the object detector \cite{fastercnn} at each frame. 
In detail, the frames are sampled at regular intervals to represent the video clip in a condensed form and in that way the computational complexity is reduced. The Conv3D layer is passed through the sampled clip to extract the feature information in patch representations following the procedure of \cite{motionformer,video_swin}.  The object detector \cite{fastercnn} pre-trained on Ego4D dataset \cite{ego4d} is used to identify and locate objects in $V$. The aforementioned information includes the location of the objects in terms of the bounding boxes ($x$, $y$, $w$, $h$ referring to center coordinates, width, and height of the bounding box, respectively). Following that, object embeddings are obtained by passing the object detection results, represented in terms of bounding boxes, through a Multi-Layer Perceptron (MLP) (please see Fig. \ref{fig:method}). \\

\noindent
\textbf{Object Guided Attention.}
We use Objects-Guided Multi-head Attention to efficiently fuse spatiotemporal information across the video clip, and object detections and then infer long-term dependencies across both. Using a single attention head does not suffice as our goal is to allow detection embeddings to attend to co-related patches from the video clip. Therefore, we modify the Multi-Head Attention described in \cite{vaswani_attn} in a way that it can take the inputs from both modalities. To do so, we set Query $Q$, Key $K$, and Value $V$ as follows.

\begin{align}
    \begin{array}{c}
     Q = f_{vid}(F_i), \text{where}\;i \in [1,..N], \\
     \\
    K, V = f_{obj}(O_j), \text{where}\;j \in [1,...M],\\
    \\
    \text{Object-Guided Attention(Q,K,V)} = Concat(h_1, ... h_h)W_o, \\
    \text{where}\; h_i = Attention({QW_i}^Q, {KW_i}^K, {VW_i}^V), \\
    \\
    \text{and Attention(Q, K, V)} = softmax(\dfrac{{{QK}^T}}{d_k})V  
    \\
    \end{array}  
\end{align}
where ${W_i}^Q, {W_i}^K, and {W_i}^V$ are learnable parameter matrices and $d_k$ represents the dimensions of $K$. The output of this Object-Guided Multi-Attention is the attended patch embeddings for the provided object features, denoted as $F_i$. \\

\noindent
\textbf{Transformer.} It is important to model interactions across frames to better contextualize the motion information and understand the object interaction that will happen in the future. For this purpose, we use a transformer that takes as inputs the attentive patch tokens and the object queries from the last observed frame and then produces the prediction results (shown as next active object (NAO) outputs in Fig. \ref{fig:method}) for each associated query. \\

\vspace{-8pt}
\noindent
\textbf{Encoder.} 
We feed $F_i$ to the encoder of our transformer network with Multi-Scale Attention blocks \cite{memvit2022} for better temporal support across all the frames. Similar to \cite{hand_obj_joint,detr,vaswani_attn}, we include spatial and temporal position encodings to the patch tokens $F_i$. Spatial and temporal position encoding allows for a sequence representation of patches and information is passed along the temporal dimension. In the end, we use the output from the last layer of the encoder: $F_e$ to be passed to the decoder.  \\

\vspace{-8pt}
\noindent
\textbf{Decoder.} The output from the encoder: $F_e$ represents the overall extracted high-level information associated with a video clip. The decoder aims to decode that information for each object query in the last observed frame. 
Object queries are formulated as the Regions of Interest (ROIs) extracted using the object detections obtained in the feature extraction step from the last observed frame. This implementation is chosen since the next active object has to be identified in this particular frame. If there are fewer detections for a given clip, then we append learnable queries to the query set. The combined features are then sent to another MultiScale attention block  to produce the predictions regarding the next active object in terms of noun class ($\hat{n}$), bounding box ($\hat{b}$), the verb depicting the future action ($\hat{v}$), and the time to contact (TTC) ($\delta$) corresponding to each object related to the queries. \\

\vspace{-8pt}
\noindent
\textbf{Learning.}
In order to train the proposed model, two different types of loss functions are employed: (1) classification loss and (2) regression loss. 
For classification purposes, the model is trained to predict the future action, represented by $\hat{v}$, and the label of the next active object, represented by $\hat{n}$. To achieve this, the model uses a cross-entropy loss, which measures the difference between the predicted and ground-truth labels. Cross-entropy loss is a commonly used loss function for classification tasks, and suits to purpose since it is able to handle the multiple class classification. The regression task includes the prediction of the bounding box of the next active object, represented by $\hat{b}$, and the TTC, represented by $\delta$. The model uses a combination of Mean Square Error (MSE) loss and Smooth L1 loss \cite{smoothl1}.
The final loss is the combination of each loss aggregated as:
\begin{equation}
    \mathcal{L} = \lambda_{1}\mathcal{L}_{box} + \lambda_{2}\mathcal{L}_{noun} + \lambda_{3}\mathcal{L}_{verb} + \lambda_{4}\mathcal{L}_{ttc}
    \label{eq_overall}
\end{equation}
where $\lambda_{1}, \lambda_{2}, \lambda_{3}, \lambda_{4} \in {R}$ are hyperparameters.
Notice that these loss functions provide a comprehensive way to train the model for action anticipation, taking into account both the classification and regression aspects of the task, allowing GANO to predict both the class of the next active object, the bounding box, the action that will be performed, and the time to contact of the action.

\begin{table*}[t]
\caption{Results of the proposed method (Ours) and the SOTA methods for different output targets: bounding box ($\hat{b}$), next active object class label ($\hat{n}$), future action ($\hat{v}$), and the time to contact ($\delta$) based on their Average Precision ($AP$). See text for the explanation of ``\emph{additional bbox + noun}". The best results and the second-best results with respect to others of each column are shown in \textbf{bold} and \underline{underlined}.}
\resizebox{\linewidth}{!}{
\begin{tabular}{|l|cccccccc|} \hline
Models & $AP_{\hat{b}}$ & $AP_{\hat{b} + \hat{n}}$ & $AP_{\hat{b} + \hat{n} + \delta}$ & $AP_{\hat{b} + \hat{n} + \hat{v}}$ & $AP_{\hat{b} + \hat{n} + \hat{v} + \delta}$ & $AP_{\hat{b} +  \delta}$ & $AP_{\hat{b} + \hat{v}}$ & $AP_{\hat{b} + \hat{v} + \delta}$  \\  \hline
Slowfast \cite{feichtenhofer2019slowfast,ego4d} & 40.5 & 24.5 & 5.0 & 0.3 & 0.06 & 8.16 & 0.34 & 0.06  \\

Slowfast \cite{ego4d} (w/ Transformer) & \underline{40.5} & \underline{24.5} & 4.5 & 4.37 & 0.73 & 7.5 & 8.2 & 1.3 \\ 

AVT \cite{avt} & \underline{40.5} & \underline{24.5} & 4.39 & 4.52 & 0.71 & 7.12 & 8.45 & 1.15 \\

ANACTO \cite{anacto} & \underline{40.5} & \underline{24.5} & 4.55 & 5.1 & 0.91 & 7.47 & 8.9 & 1.54 \\

MeMVIT \cite{memvit2022} & \underline{40.5} & \underline{24.5} & 4.95 & \underline{5.89} & \underline{1.34} & 9.27 & \underline{10.04} & 2.11 \\ \hline

\textbf{Ours} w/o guided attention & \underline{40.5} & \underline{24.5} & 4.2 & 4.22 & 0.75 & 9.01 & 7.1 &  1.22 \\

\textbf{Ours} (w/ guided attention) & \underline{40.5} & \underline{24.5} & \textbf{5.9} & \textbf{6.2} & \textbf{1.7} & \underline{10.1} & \textbf{10.56} & \textbf{2.77} \\ 

\textbf{Ours} (additional bbox + noun) & \textbf{45.2} & \textbf{25.8} & \underline{5.05} & 5.6 & 1.2 & \textbf{11.2} & 9.7 & \underline{2.29} \\ \hline

\end{tabular}}
\label{table:results}
\vspace{-7pt}
\end{table*}

\begin{figure*}[t!]
\centering
\includegraphics[width=\linewidth, height=0.1\linewidth]{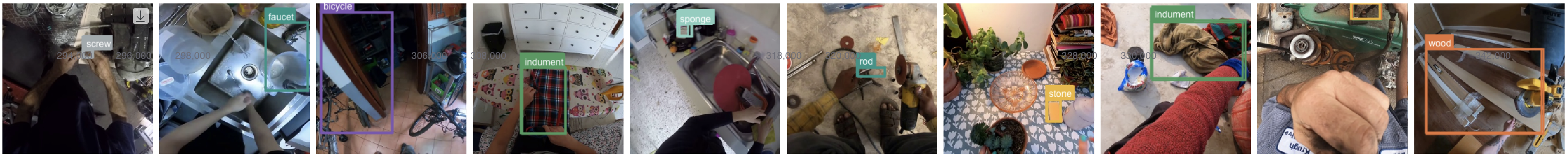}
\includegraphics[width=\linewidth, height=0.1\linewidth]{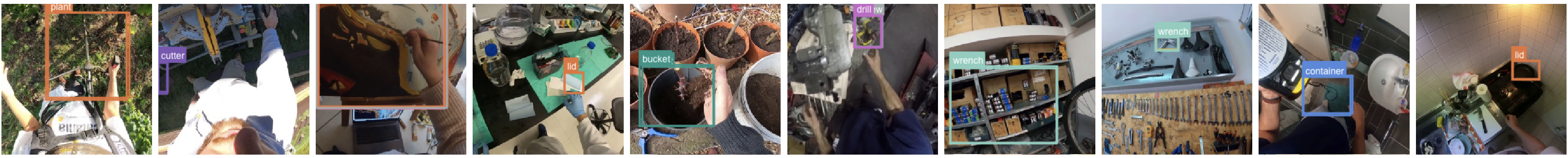}
\includegraphics[width=\linewidth, height=0.1\linewidth]{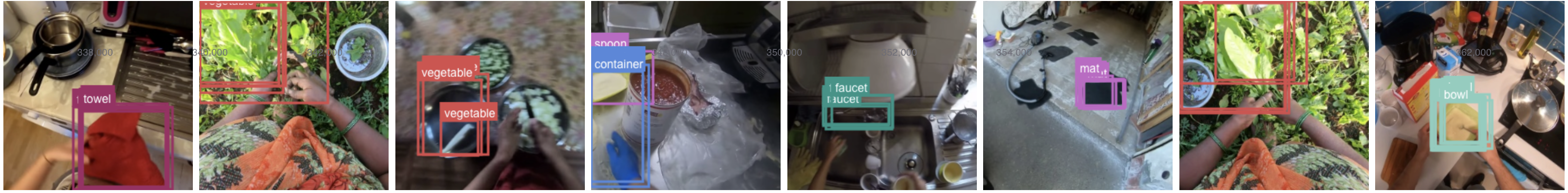}
\includegraphics[width=\linewidth, height=0.1\linewidth]{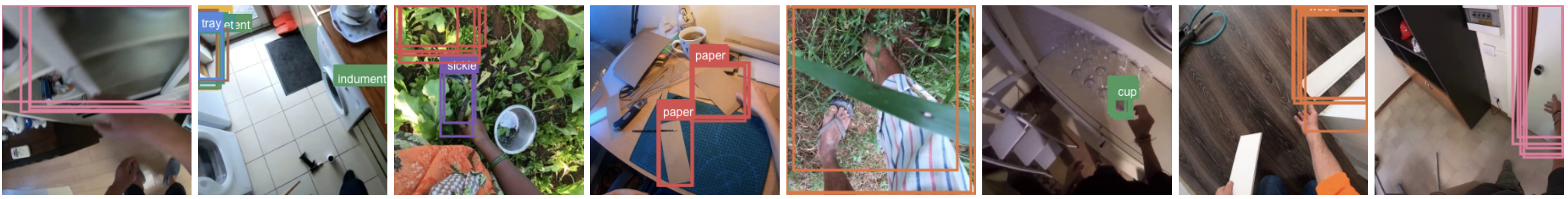}
\caption{Qualitative results of our model (additional bounding boxes and nouns are given) on Ego4D dataset \cite{ego4d} when predicting the bounding box and corresponding label for the next active object in the last observed frame.
}
\centering
\label{fig:qualitative}
\vspace{-10pt}
\end{figure*}

\section{Experiments}
This section describes the experimental setup, implementation details, and the comparative evaluation of our method against the state-of-the-art methods on the Ego4D dataset \cite{ego4d}.

\noindent
\textbf{Ego4D} is the largest first-person dataset, which has recently been released. The dataset was split into five different categories, each focusing on a different task, combining a total of 3,670 hours of egocentric videos across 74 locations. In this paper, we focus on the forecasting split (i.e., we used the corresponding training and testing splits as supplied), containing more than 1000 videos for a total of 960 hours, annotated at 30 fps for the STA task. The annotations are for the next active objects in the last observed frame. This is the only dataset that publicly provides annotations for the next active objects for the aforementioned task. Given that the aim is to predict the noun class ($\hat{n}$), bounding box ($\hat{b}$), 
the verb depicting the future action ($\hat{v}$), and the TTC ($\delta$) for a given video clip, we compare our method with the following \textbf{state-of-the-art (SOTA)} action anticipation: AVT \cite{avt}, MeMViT \cite{memvit2022}, and STA: Slowfast \cite{feichtenhofer2019slowfast,ego4d} (both the CNN and the transformer backbone was utilized) methods. The evaluation is based on comparing the predictions of the verb category and the time to contact, using the object detection bounding boxes and the corresponding noun labels from a pre-trained object detector \cite{fastercnn}. This allows us to thoroughly evaluate our method to show whether it can improve the performance of existing action anticipation methods. To ensure a fair comparison, we train each of the SOTA using the configurations specified in their original papers to predict the \emph{future action} and the \emph{time to contact}. The results of SlowFast (with CNN-based and a Transformer-based backbone) \cite{feichtenhofer2019slowfast} were obtained by using the implementation described in the Ego4D \cite{ego4d} paper, which is specifically tailored for the STA task. \\

\noindent
\textbf{Implementation details. }
The encoder of the proposed method is based on the implementation of MViT \cite{mvit} (16 layers) with 16 input frames at a sampling rate of 4. We used the encoder which is pre-trained on Kinetic-400 \cite{kinetic_400}. We apply random horizontal flipping and random cropping of size 224 from frames resized such that the short side $\in [256, 340]$ as data augmentation. GANO was trained with an SGD optimizer for 30 epochs with a cosine learning rate of $1e-5$ with a batch size of 4 and a weight decay of $1e-6$ on two NVIDIA-SMI Tesla V100 GPU. We kept the values of $\lambda_2$ as 1.0 and $\lambda_1$ as 0.5 (see Eq. \ref{eq_overall}) respectively, during the training of GANO. \\

\noindent
\textbf{Evaluation Metric.} 
In our study, we use the evaluation metrics proposed in \cite{ego4d} to assess the performance of our method for the STA task. The metrics comprise the Average Precision of four different combinations of the next active object-related predictions: noun class ($\hat{n}$), bounding box ($\hat{b}$), future action ($\hat{v}$), and TTC ($\delta$). The top-1 accuracy is used to evaluate the performance of future action ($\hat{v}$) and NAO label ($\hat{n}$) predictions. For bounding boxes ($\hat{b}$) and time to contact ($\delta$), the predictions are considered correct if the predicted boxes have an Intersection over Union (IoU) value greater than or equal to 0.5 and the absolute difference between the predicted and ground-truth time to contact is less than or equal to 0.25 seconds ($|\hat{y}{ttc} - y{ttc}| \leq 0.25$). In the case of combined predictions involving two or more unknowns, the prediction is deemed correct only if all the unknowns are predicted correctly.  \\

\vspace{-8pt}
\noindent
\textbf{Results.}
The results are reported in Table \ref{table:results}. In that table, for all the models, except the last row, the models are trained to predict only the future action ($\hat{v}$) and TTC ($\delta$). For all methods, the object detector \cite{fastercnn} was used for the prediction of location ($\hat{b}$) and the noun label ($\hat{n}$) of the next active object (therefore results: $AP_{\hat{b}}$ and $AP_{\hat{b} + \hat{n}}$ off all methods are the same). The results demonstrate that GANO outperforms all baseline methods across all metrics evaluated. We also conducted an ablation study to investigate the impact of the \emph{guided attention} mechanism. The proposed method, which does not use guided attention, involves the fusion of object features with patch features through concatenation prior to feeding them into the transformer (shown as w/o guided attention in Table \ref{table:results}). As seen, in that case, there is a drastic drop in performance compared to using guided attention. In other words, fusing objects and spatiotemporal features with concatenation is not a favorable fusion method. Instead, we show that \emph{guided attention} remarkably contributes to the performance of our model (shown as w/ guided attention in Table \ref{table:results}).

We also evaluate the performance of the proposed method in order to predict bounding boxes and noun labels, in other words without simply relying on the object detection \cite{fastercnn} (the corresponding results are given in the last row of Table \ref{table:results}). To do this, we append learnable object tokens with ROIs embeddings to compute a fixed set of 50 object queries. On the one hand, such implementation results in a much-improved performance in detecting the next active objects' location ($\hat{b}$) and class label ($\hat{n}$). However, this leads to a slight drop in anticipating the motion-related outputs (i.e., the evaluations including verb predictions $\hat{v}$). We additionally visualize the qualitative results of the aforementioned version of GANO for the predictions of bounding boxes ($\hat{b}$) and NAO class label ($\hat{n}$) in Fig. \ref{fig:qualitative}. As seen GANO trained by appending learnable object tokens with ROIs embeddings is good at detecting various types of objects in terms of their class labels and is also precise to detect the corresponding bounding boxes.
\vspace{0.5cm}

\section{Conclusion}
\label{sec:conclusion}
In this paper, we have presented a novel approach for the next active object-based action anticipation in egocentric videos using a Guided Attention mechanism. Our method achieved state-of-the-art results on the largest egocentric dataset that supplies the relevant annotations publicly. We show that the guided attention mechanism is effective to learning from the object features and the spatio-temporal features simultaneously, such that it results in a noticeable performance improvement with respect to other types of fusions. Furthermore, we demonstrate that it is possible to improve the proposed method's next active object location and class label prediction performances by simply appending learnable object tokens with region of interest embeddings. Future work will investigate the usage of the proposed method for video summarization and human-robot interaction analysis.

\newpage
\bibliographystyle{IEEEbib}
\bibliography{main}

\end{document}